\def\year{2020}\relax
\documentclass[letterpaper]{article} 
\usepackage{aaai20}  

\usepackage{multirow}
\usepackage{amsmath}
\usepackage{dsfont}

\usepackage{times}  
\usepackage{helvet} 
\usepackage{courier}  
\usepackage[hyphens]{url}  
\usepackage{graphicx} 
\usepackage{subfig}
\usepackage{graphicx}  
\usepackage{hyperref}

\usepackage{color}
\usepackage{xspace}
\usepackage{url}
\usepackage[subtle]{savetrees} 

\usepackage{wrapfig}

\usepackage{booktabs} 

\usepackage{array}
\newcommand{\PreserveBackslash}[1]{\let\temp=\\#1\let\\=\temp}
\newcolumntype{C}[1]{>{\PreserveBackslash\centering}p{#1}}
\newcolumntype{R}[1]{>{\PreserveBackslash\raggedleft}p{#1}}
\newcolumntype{L}[1]{>{\PreserveBackslash\raggedright}p{#1}}
\newcolumntype{M}[1]{>{\centering\arraybackslash}m{#1}}

\newif\ifsubmit
\submitfalse

\ifsubmit
    \newcommand{\amir}[1]{}
        \newcommand{\kevin}[1]{}
    \newcommand{\pratik}[1]{}
    
    \newcommand{\todo}[1]{}
\else
    \newcommand{\amir}[1]{\textcolor{blue}{Amir: #1}}
    \newcommand{\kevin}[1]{\textcolor{green}{Kevin: #1}}
    \newcommand{\pratik}[1]{\textcolor{cyan}{Pratik: #1}}
    
    \newcommand{\todo}[1]{\textcolor{red}{TODO: #1}}
\fi

\newcommand{\paratitle}[1]{\noindent\textbf{\textit{#1.}}\xspace}

\newcommand{\ie}[0]{\emph{i.e.,}\xspace}
\newcommand{\etal}[0]{\emph{et al.}\xspace}
\newcommand{\eg}[0]{\emph{e.g.,}\xspace}

\usepackage{cleveref} 

\crefformat{section}{\S#2#1#3} 
\crefformat{subsection}{\S#2#1#3}
\crefformat{subsubsection}{\S#2#1#3}

\title{Can Attention Masks Improve Adversarial Robustness?}

\author{\Large \textbf{
   Pratik Vaishnavi\textsuperscript{\rm 1},
   Tianji Cong\textsuperscript{\rm 2},
    Kevin Eykholt\textsuperscript{\rm 2},
    Atul Prakash\textsuperscript{\rm 2},
    Amir Rahmati\textsuperscript{\rm 1}
    } \\ 
\textsuperscript{\rm 1}Stony Brook University ~~~~~ 
\textsuperscript{\rm 2}University of Michigan\\ 
pvaishnavi@cs.stonybrook.edu, \{congtj,keykholt,aprakash\}@umich.edu, amir@rahmati.com}

\begin{document}

\maketitle

\begin{abstract}
Deep Neural Networks (DNNs) are known to be susceptible to adversarial examples. Adversarial examples are maliciously crafted inputs that are designed to fool a model, but appear normal to human beings.  Recent work has shown that pixel discretization can be used to make classifiers for MNIST highly robust to adversarial examples. However, pixel discretization fails to provide significant protection on more complex datasets. In this paper, we take the first step towards reconciling these contrary findings. Focusing on the observation that discrete pixelization in MNIST makes the background completely black and foreground completely white, we hypothesize that the important property for increasing robustness is the elimination of image background using attention masks before classifying an object. To examine this hypothesis, we create foreground attention masks for two different datasets, GTSRB and MS-COCO. Our initial results suggest that using attention mask leads to improved robustness. On the adversarially trained classifiers, we see an adversarial robustness increase of over 20\% on MS-COCO.
\end{abstract}

\section{Introduction}
Deep Neural Networks are employed in a wide range of applications ranging from autonomous systems to trading and healthcare. This has resulted in an increased attention to their security. One of the primary focuses of these efforts has been defense against adversarial examples. Adversarial examples~\cite{szegedy2014intriguing} can be generated by adding carefully-crafted imperceptible noise to a normal input example. Such an adversarial example can be used to trigger a misclassification on a target model for image classification tasks (\eg a road-sign classifier in a self-driving car). Many techniques have been developed to tackle this problem ~\cite{xie2019feature,madry2018towards,lamb2018fortified}, one of the popular one being adversarial training.


In analyzing an adversarially trained DNN on MNIST, Madry~\etal~\cite{madry2018towards} found that the first layer filters turned out to be  thresholding-filters that were acting as a de-noiser for the grayscale MNIST images. Follow up experiments by Schott~\etal~\cite{schott2018towards} showed that training a DNN with binarized MNIST images (where each pixel was discretized to be either made completely black or completely white using a static threshold) resulted in significantly improved adversarial accuracy without any adversarial training or negative impact on normal performace. In other words, a pipeline that first thresholds each pixel in an MNIST image to 1 or 0 and then classifies the resulting image with a naturally trained model has a very high degree of adversarial robustness, without requiring any adversarial training. Subsequent works, however, found that a simple binarization was not effective for more complex datasets such as CIFAR-10~\cite{discretepixelization}. 

\begin{figure}[t]
\includegraphics[width=\columnwidth]{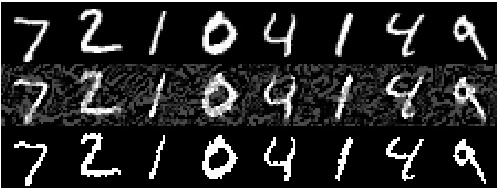}
\caption{Visualizing images from the MNIST dataset. The first row contains natural images, the second row contains corresponding adversarial images and the third row contains binarized adversarial images (threshold = $0.5$). Binarization removes almost all the adversarial noise.}
\label{fig:mnist}
\end{figure}


In contrast to previous work, we observe that binarization on MNIST acts as an approximation for the process of \textit{foreground-background} separation. Figure~\ref{fig:mnist} presents a sample image from the MNIST dataset. MNIST dataset consist of $28\times 28$ pixel images of handwritten white digits on a black background and are among the simplest datasets in ML. Adversarial attacks on the MNIST typically do not distinguish between the digits and the background and may similarly manipulate both to achieve their goal. When examining the adversarial noise in MNIST adversarial examples for a naturally-trained model, we see that the adversarial noise is spread throughout the image, including in  background regions of the image that we might not consider to carry any predictive signals. Binarization on MNIST as a pre-processing step snaps most of the background pixels back to the original value on which the model was trained (\ie $0$), hence reducing the attack surface.  

Therefore, it may be more accurate to characterize binarization on MNIST as foreground-background separation rather than simple pixel discretization for more complex image datasets. If the hypothesis is true, then we should see improved robustness on other datasets by simply separating the background from the foreground and masking the background prior to training and classification, i.e., applying a foreground-attention mask on the dataset.

Towards validating the above hypothesis, given a classifier and a dataset, we introduce an additional pre-processing step where a foreground attention mask is applied to the model's input before classification. A challenge in testing our hypothesis is determining the foreground attention mask. Unfortunately, most image datasets on which adversarial testing is done (e.g., CIFAR-10, ImageNet) lack sufficient ground truth data for foreground attention masks.  To address the challenge, we generate two datasets with foreground attention masks from existing datasets:  The German Traffic Sign Recognition (GTSRB)~\cite{Stallkamp2012} and MS-COCO~\cite{lin2014microsoft}. For the GTSRB dataset, we took advantage of the typical color distribution in images and that a road sign often lies in the center of the image,  and used them to design a custom attention mask generator by doing random sampling of pixels near the center of the image along with a min cut-max flow algorithm to create the foreground attention mask. For the MS-COCO dataset, we use the segmentation masks included with the dataset to create a cropped image of the object of interest and its foreground mask. 

Our preliminary results suggest that a classification pipeline that utilizes foreground attention masks experiences improved adversarial robustness with, at worst, no impact on natural accuracy. On the naturally trained classifiers, the adversarial accuracy improves by 0.73\% on MS-COCO and around 19.69\% on GTSRB. Robustness improvements were also found on combining usage of attention masks with adversarial training. Thus, this paper makes the following contributions: 
\begin{itemize} 
\item We create two datasets based on the GTSRB and MS-COCO that allow exploration of attention mask effects on adversarial examples. These datasets are available at \href{https://drive.google.com/drive/folders/1CtYutfdN6cvC2cLmhz-nJeuLkLMMhd78?usp=sharing}{[link]}.

\item We take the first step toward exploring the effect of attention masks on improving model robustness in image classification tasks. We show that attention masks have certain effect on improving adversarial performance against PGD adversary.

\end{itemize}

\section{Background}

\paratitle{Discrete Pixelization Defenses}
Developing adversarial defenses towards robust classification has received significant attention in recent years~\cite{madry2018towards,Liao2018DefenseAA}. Among these, defense methods that  pre-process inputs to improve robustness are potentially attractive because the pre-processed input can be passed to existing classifiers for improved robustness.
Unfortunately, some of these methods were vulnerable to stronger adaptive adversarial attacks~\cite{athalye2018obfuscated}, raising doubts on the effectiveness of pre-processing strategies.

One pre-processing strategy that has stood attacks well against stronger adaptive adversarial attacks is that of binarization for the MNIST dataset. Unfortunately, binarization, which converts each pixel value to fewer bits did not provide a significant benefit on more complex datasets and, in some cases, negatively impacted test accuracy~\cite{discretepixelization}. Chen \etal provide theoretical insights into the phenomenon, concluding that discrete pixelization is unlikely to provide significant benefit.

\paratitle{Semantic Segmentation}
Semantic segmentation of images has applications in diverse fields like biomedical imaging \cite{ronneberger2015u}. \cite{wu2019fastfcn,chen2017rethinking} describe semantic segmentation techniques for complex real-word datasets like MSCOCO and Cityscapes. However, \cite{arnab2018robustness} have shown that DNN-based segmentation pipelines can  be susceptible to adversarial attacks, though other work has shown that such attacks may be successfully detected~\cite{xiao2018characterizing}, potentially providing a  defense. 
We note that, unlike our work, prior work on robustness of semantic segmentation looked at robustness of segmentation model itself and not the impact of foreground-background separation on robustness of classification of a foreground object.

\paratitle{Attention masks}
According to work by Xie \etal on feature denoising, they discovered that adversarial noise causes machine learning models to focus on semantically uninformative information in the input, whereas the opposite is true for natural clean inputs. Thus, rather than relying on the model to identify relevant input features, we explore if we can force the network to focus on important portions of the image (\eg the foreground object). Harley \etal~\cite{Harley_2017_ICCV_attention} proposed the idea of segmentation-aware convolution networks that rely on local attention masks, which are generated based on color-distance among neighboring pixels, and found that it can improve semantic segmentation and optical flow estimation. Our work aims to understand if  attention masks can also be useful for improvement of robustness of image classification.
\section{Our Approach}
In this work, our goal is to examine if isolating predictive signals in the form of foreground features has benefits in terms of adversarial robustness while having minimal impact on model performance on natural inputs. We examine, using two datasets, that training a model on foreground pixels helps it perform well not only on natural images, but makes it robust against adversarial images as well.

Let $\mathbf{X}$ be a set of images drawn from a distribution. Let's consider the task of image classification defined on $\mathbf{X}$. Traditionally, in image classification, we restrict each image to contain only one object. An image $x^{(i)} \in \mathbf{X}$ can be divided into foreground and background pixels. By definition, foreground pixels are pixels that are a part of the object and every other pixel can be considered as a part of the background. In this paper we make the assumption that the foreground pixels, on their own, carry sufficient predictive power for the task of image classification. Additionally, removing background pixels restricts the input space that the adversary can attack, inhibiting its ability to trigger misclassification in the target model.

For an image $x^{(i)} \in \mathbf{X}$ of resolution $m \times n$, let's define a foreground image $x^{(i)}_{FG} = F(x^{(i)})$, where

\begin{equation*}
    F(x^{(i)}) = \begin{cases}
    x^{(i)}_{jk} &\text{if $x^{(i)}_{jk} \in S^{(i)}_{FG}$}\\
    0 &\text{else}
    \end{cases}
    \quad j = 1 \dots m, k = 1 \dots n
\end{equation*}

where, $S^{(i)}_{FG}$ is the set of foreground pixels for image $x^{(i)}$.  We generate $\mathbf{X}_{FG}$ containing foreground images $x^{(i)}_{FG}$ $\forall x^{(i)} \in \mathbf{X}$. Based on the above, we evaluate two class of models: (1) model trained on $\mathbf{X}$, (2) model trained on $\mathbf{X}_{FG}$. For both, we perform Natural (N) and Adversarial (A) training.


We assume access to foreground masks in our experiments. Thus, our work provides an upper bound on the potential benefit that can be provided by foreground attention masks on model robustness, assuming foreground attention masks can be robustly found. There is some potential hope since recent work has shown that adversarial attacks may be successfully detected on segmentation algorithms~\cite{xiao2018characterizing} using statistical tests, potentially providing a basis for a defense. 

\section{Dataset Creation with Foreground Masks}
\paratitle{MS-COCO} We pre-process the MS-COCO dataset to make it compatible with the task of image-classification. Particularly, we use the following pre-processing steps:

\begin{itemize}
    \item We make use of the semantic segmentation masks and object bounding box annotations to generate image, mask, label pairs such that each image contains object(s) of one label only, or in other words, the mask corresponding to an image contains annotations for one object only.
    \item To deal with objects having overlapping bounding box regions, we explicitly black-out pixels corresponding to the extra objects.
    \item We adjust the crop dimensions in order to extract square image patches, and resize all the extracted image patches to $32\times32$.
    \item Due to high class-imbalance in the resultant dataset, we ignore the person class (over frequent) and short-list top 10 classes from the remaining classes based on the frequency.
\end{itemize}

We call this modified MS-COCO dataset as \textbf{MS-COCO-IC}. Table \ref{tab:mscoco-data-stats} shows the statistics for this dataset. The images in this dataset contain $\approx 56\%$ foreground pixels. Figure~\ref{fig:mscoco-cropping} gives an example of an original image, the cropped image, and the final image that is used for training a classifier.

\begin{table}[hbt!]
\begin{center}
\begin{tabular}{ @{}ccc@{} }
 \toprule
\multirow{2}{*}{\textbf{Class}} & \multicolumn{2}{c}{\textbf{Number of Images}} \\ \cmidrule(lr){2-3}
 & \textbf{Train} & \textbf{Test} \\ 
\hline
Chair & 21674 & 11077 \\
Car & 18498 & 9594 \\
Book & 12094 & 6188 \\
Bottle & 10923 & 5735 \\
Dinning table & 10291 & 5274 \\
Umbrella & 6510 & 3309 \\
Boat & 5531 & 2797 \\
Motorcycle & 5340 & 2703 \\
Sheep & 4748 & 2432 \\
Cow & 4391 & 2162\\
\midrule
Total & 100000 & 51271 \\
\bottomrule
\end{tabular}
\caption{Number of images per class in the train and test set of our MSCOCO-IC dataset.}
\label{tab:mscoco-data-stats}
\end{center}
\vspace{-0.2in}
\end{table}

\paratitle{GTSRB} The German Traffic Sign Recognition Benchmark (GTSRB) is a dataset containing $50,000$ color images of 43 different classes of road signs, with high class imbalance. Images in GTSRB model various viewpoints and light conditions. We use a customized segmenter based on the graph cut algorithm to obtain foreground masks. One favorable aspect of GTSRB is that majority of the traffic signs are centrally located in the image, have regular shapes, and usually possess a sharp color in contrast to the background. These features match the assumptions of our customized segmenter. The images in this dataset contain $\approx 25\%$ foreground pixels. We associate the low percentage of foreground pixels to the imperfections of our ad hoc segmenter.
Figure~\ref{fig:gtsrb-cropping} gives an example of an image, computed mask by our segmenter, and the final image that is used for training a classifier.

\begin{table}[tb!]
    \centering
    \resizebox{0.88\columnwidth}{!}{%
    \begin{tabular}{@{}M{25mm}M{25mm}M{25mm}@{}}
    \includegraphics[width=0.15\textwidth]{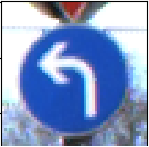} &     \includegraphics[width=0.15\textwidth]{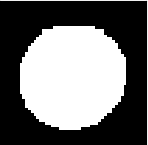} &   \includegraphics[width=0.15\textwidth]{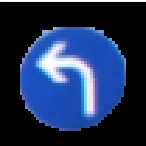}
    \end{tabular}
    }
\caption{Visualizing examples in the GTSRB-IC dataset. We display the natural image, foreground mask and the foreground masked image from left to right.}
\label{fig:gtsrb-cropping}
\end{table}

\begin{table}[tb!]
    \centering
    \resizebox{0.88\columnwidth}{!}{%
    \begin{tabular}{@{}M{40mm}M{25mm}M{25mm}@{}}
    \includegraphics[width=0.23\textwidth]{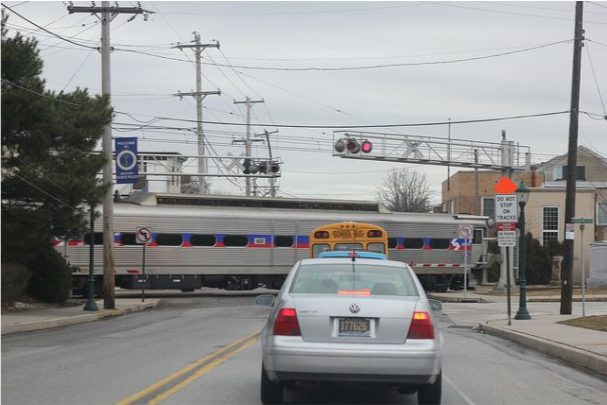} &     \includegraphics[width=0.15\textwidth]{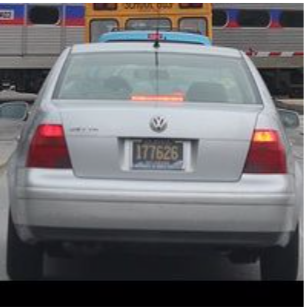} &   \includegraphics[width=0.15\textwidth]{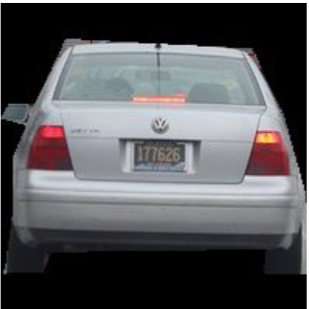}
    \end{tabular}
    }
\caption{Visualizing examples in the MS-COCO-IC dataset. We display the original image from the MS-COCO dataset, cropped image of object of interest, and the foreground masked image from left to right.}
\label{fig:mscoco-cropping}
\end{table}


\section{Results}
For our experiments, we train two set of models: (1) on natural images; (2) on foreground masked image, both naturally and adversarially. For both, we use the VGG-19 classifier. We evaluate these models against a $10$-step $\mathbf{L}_{\infty}$ bounded PGD adversary with step size of $\frac{2}{255}$ and $\epsilon = \frac{8}{255}$. Treating the performance of models trained on $\mathbf{X}$ as a baseline, we calculate potential gains in robustness in the models trained on $\mathbf{X}_{FG}$. We repeat the above experiments for both GTSRB-IC and MS-COCO-IC datasets. Note that in the case of $\mathbf{X}_{FG}$ models, the adversary is only given access to the foreground pixels. We summarize our results in Table ~\ref{tab:results}.

\vspace{0.1in}
\paratitle{MS-COCO-IC}
We can observe from the results that both the naturally trained classifiers exhibit comparable vulnerability to PGD adversary. However, in case of the adversarially trained classifiers, natural and adversarial accuracy increases by $11.41\%$ and $22.82\%$ respectively, on using foreground attention masks. This validates our hypothesis that foreground attention has a positive effect on a model's classification performance and robustness.

\begin{table}[hbt!]
\begin{center}
\resizebox{0.95\columnwidth}{!}{%
\begin{tabular}{ @{}cccccc@{} }
\toprule
\multirow{2}{*}{\textbf{Data}} & \multirow{2}{*}{\textbf{Training}} & \multicolumn{2}{c}{\textbf{MSCOCO-IC}} & \multicolumn{2}{c}{\textbf{GTSRB-IC}} \\ \cmidrule(lr){3-4} \cmidrule(lr){5-6}
& & \textbf{Natural} & \textbf{PGD} & \textbf{Natural} & \textbf{PGD} \\ 
\midrule
$\mathbf{X}$ & N & 79.46\% & 2.28\% & 98.04\% & 18.69\% \\
$\mathbf{X}_{FG}$ & N & \textbf{81.64\%} & \textbf{3.01\%} & 98.04\% & \textbf{38.38}\% \\
\midrule
$\mathbf{X}$ & A & 61.51\% & 30.80\% & 89.54\% & 55.25\% \\
$\mathbf{X}_{FG}$ & A & \textbf{72.92\%} & \textbf{53.62}\% & \textbf{91.20\%} & \textbf{64.57}\% \\
\bottomrule
\end{tabular}
}
\caption{Comparing adversarial robustness of models trained on natural images versus foreground masked images. For both datasets. we observe increased robustness against a PGD adversary when the model and the adversary have access to foreground features only.}
\label{tab:results}
\end{center}
\vspace{-0.2in}
\end{table}

\paratitle{GTSRB-IC}
Similar to the previous set of results, we see that the model adversarially trained using $\mathbf{X}_{FG}$ is more robust to a PGD adversary than a model adversarially trained using $\mathbf{X}$ ($+9.32\%$). Additionally, we see that the model naturally trained on $\mathbf{X}_{FG}$ exhibits considerable improvement in robustness as compared to the model naturally trained on $\mathbf{X}$ ($+19.69\%$).



\section{Conclusion}
We study the use of foreground attention masks for improving the robustness of deep neural networks against $L_{\infty}$-bounded PGD attack. We develop two new datasets based on MS-COCO and GTSRB, to examine these effects. Our preliminary results suggest positive effects in using foreground masks for improving adversarial robustness against PGD adversary. For an adversarially trained model on the MS-COCO-IC dataset, foreground attention masks improved adversarial accuracy by $21.8\%$. For the GTSRB-IC dataset, when adversarially training a model with foreground masked images, we observe a smaller improvement of $1.7\%$ in adversarial accuracy. Initial results are promising however, further work must be done to verify the effect of foreground attention masks on adversarial robustness and to develop an reliable method to extract these foreground masks automatically.

Prior work by Simon-Gabriel \etal ~\cite{simon2018adversarial} suggest that our masking technique improves adversarial robustness due to a reduction in the number of input features. Against a first-order adversary (\eg PGD attack), they establish that the adversarial robustness scales as $1/\sqrt{d}$ where $d$ is the number of input features. Therefore, masking some of the input features should improve adversarial robustness to some degree. For future work, we intend to investigate this relationship further as we believe that the relative importance of a feature for classification may suggest additional robustness. Also, this relationship might help better explain certain trends that we observe in the results. Such work will help better understand the usefulness of foreground masks in context of adversarial robustness against first-order adversaries.


\bibliographystyle{aaai}
\bibliography{references}

\begin{thebibliography}{}

\bibitem[\protect\citeauthoryear{Arnab, Miksik, and
  Torr}{2018}]{arnab2018robustness}
Arnab, A.; Miksik, O.; and Torr, P.~H.
\newblock 2018.
\newblock On the robustness of semantic segmentation models to adversarial
  attacks.
\newblock In {\em IEEE Conf. on Computer Vision and Pattern Recognition}.

\bibitem[\protect\citeauthoryear{Athalye, Carlini, and
  Wagner}{2018}]{athalye2018obfuscated}
Athalye, A.; Carlini, N.; and Wagner, D.
\newblock 2018.
\newblock Obfuscated gradients give a false sense of security: Circumventing
  defenses to adversarial examples.
\newblock In {\em Intl. Conf. on Machine Learning}.

\bibitem[\protect\citeauthoryear{Chen \bgroup et al\mbox.\egroup
  }{2017}]{chen2017rethinking}
Chen, L.-C.; Papandreou, G.; Schroff, F.; and Adam, H.
\newblock 2017.
\newblock Rethinking atrous convolution for semantic image segmentation.
\newblock {\em arXiv preprint arXiv:1706.05587}.

\bibitem[\protect\citeauthoryear{Chen \bgroup et al\mbox.\egroup
  }{2018}]{discretepixelization}
Chen, J.; Wu, X.; Liang, Y.; and Jha, S.
\newblock 2018.
\newblock Improving adversarial robustness by data-specific discretization.
\newblock {\em arXiv preprint arXiv:1805.07816}.

\bibitem[\protect\citeauthoryear{Harley, Derpanis, and
  Kokkinos}{2017}]{Harley_2017_ICCV_attention}
Harley, A.~W.; Derpanis, K.~G.; and Kokkinos, I.
\newblock 2017.
\newblock Segmentation-aware convolutional networks using local attention
  masks.
\newblock In {\em Intl. Conf. on Computer Vision}.

\bibitem[\protect\citeauthoryear{Lamb \bgroup et al\mbox.\egroup
  }{2018}]{lamb2018fortified}
Lamb, A.; Binas, J.; Goyal, A.; Serdyuk, D.; Subramanian, S.; Mitliagkas, I.;
  and Bengio, Y.
\newblock 2018.
\newblock Fortified networks: Improving the robustness of deep networks by
  modeling the manifold of hidden representations.
\newblock {\em arXiv preprint arXiv:1804.02485}.

\bibitem[\protect\citeauthoryear{Liao \bgroup et al\mbox.\egroup
  }{2018}]{Liao2018DefenseAA}
Liao, F.; Liang, M.; Dong, Y.; Pang, T.; Zhu, J.; and Hu, X.
\newblock 2018.
\newblock Defense against adversarial attacks using high-level representation
  guided denoiser.
\newblock {\em IEEE Conf. on Computer Vision and Pattern Recognition}.

\bibitem[\protect\citeauthoryear{Lin \bgroup et al\mbox.\egroup
  }{2014}]{lin2014microsoft}
Lin, T.-Y.; Maire, M.; Belongie, S.; Hays, J.; Perona, P.; Ramanan, D.;
  Doll{\'a}r, P.; and Zitnick, C.~L.
\newblock 2014.
\newblock Microsoft coco: Common objects in context.
\newblock In {\em European Conference on Computer Vision}.

\bibitem[\protect\citeauthoryear{Madry \bgroup et al\mbox.\egroup
  }{2018}]{madry2018towards}
Madry, A.; Makelov, A.; Schmidt, L.; Tsipras, D.; and Vladu, A.
\newblock 2018.
\newblock Towards deep learning models resistant to adversarial attacks.
\newblock In {\em Intl. Conf. on Learning Representation}.

\bibitem[\protect\citeauthoryear{Ronneberger, Fischer, and
  Brox}{2015}]{ronneberger2015u}
Ronneberger, O.; Fischer, P.; and Brox, T.
\newblock 2015.
\newblock U-net: Convolutional networks for biomedical image segmentation.
\newblock In {\em Intl. Conf. on Medical image computing and computer-assisted
  intervention}.
\newblock Springer.

\bibitem[\protect\citeauthoryear{Schott \bgroup et al\mbox.\egroup
  }{2018}]{schott2018towards}
Schott, L.; Rauber, J.; Bethge, M.; and Brendel, W.
\newblock 2018.
\newblock Towards the first adversarially robust neural network model on mnist.
\newblock {\em arXiv preprint arXiv:1805.09190}.

\bibitem[\protect\citeauthoryear{Simon-Gabriel \bgroup et al\mbox.\egroup
  }{2018}]{simon2018adversarial}
Simon-Gabriel, C.-J.; Ollivier, Y.; Bottou, L.; Sch{\"o}lkopf, B.; and
  Lopez-Paz, D.
\newblock 2018.
\newblock Adversarial vulnerability of neural networks increases with input
  dimension.
\newblock {\em arXiv preprint arXiv:1802.01421}.

\bibitem[\protect\citeauthoryear{Stallkamp \bgroup et al\mbox.\egroup
  }{2012}]{Stallkamp2012}
Stallkamp, J.; Schlipsing, M.; Salmen, J.; and Igel, C.
\newblock 2012.
\newblock Man vs. computer: Benchmarking machine learning algorithms for
  traffic sign recognition.
\newblock {\em Neural networks}.

\bibitem[\protect\citeauthoryear{Szegedy \bgroup et al\mbox.\egroup
  }{2014}]{szegedy2014intriguing}
Szegedy, C.; Zaremba, W.; Sutskever, I.; Bruna, J.; Erhan, D.; Goodfellow, I.;
  and Fergus, R.
\newblock 2014.
\newblock Intriguing properties of neural networks.
\newblock In {\em Intl. Conf. on Learning Representations}.

\bibitem[\protect\citeauthoryear{Wu \bgroup et al\mbox.\egroup
  }{2019}]{wu2019fastfcn}
Wu, H.; Zhang, J.; Huang, K.; Liang, K.; and Yu, Y.
\newblock 2019.
\newblock Fastfcn: Rethinking dilated convolution in the backbone for semantic
  segmentation.
\newblock {\em arXiv preprint arXiv:1903.11816}.

\bibitem[\protect\citeauthoryear{Xiao \bgroup et al\mbox.\egroup
  }{2018}]{xiao2018characterizing}
Xiao, C.; Deng, R.; Li, B.; Yu, F.; Liu, M.; and Song, D.
\newblock 2018.
\newblock Characterizing adversarial examples based on spatial consistency
  information for semantic segmentation.
\newblock In {\em European Conference on Computer Vision}.

\bibitem[\protect\citeauthoryear{Xie \bgroup et al\mbox.\egroup
  }{2019}]{xie2019feature}
Xie, C.; Wu, Y.; Maaten, L. v.~d.; Yuille, A.~L.; and He, K.
\newblock 2019.
\newblock Feature denoising for improving adversarial robustness.
\newblock In {\em IEEE Conf. on Computer Vision and Pattern Recognition}.

\end{thebibliography}

\end{document}